\documentclass[sigconf]{acmart}
\AtBeginDocument{%
  }

\setcopyright{acmlicensed}
\copyrightyear{2025}
\acmYear{2025}
\acmDOI{10.1145/3746027.3758311}
\acmConference[MM '25] {Proceedings of the 33rd ACM International Conference on Multimedia}{October 27--31, 2025}{Dublin, Ireland.}
\acmISBN{979-8-4007-2035-2/2025/10}

\begin{document}

\title{StreamingCoT: A Dataset for Temporal Dynamics and Multimodal Chain-of-Thought Reasoning in Streaming VideoQA}

\author{Yuhang Hu}
\orcid{0009-0003-7771-5469}
\affiliation{
  \institution{Henan Institute of Advanced Technology, Zhengzhou University}
  \institution{Institute of Automation, CAS}
  \city{Beijing}
  \country{China}
}
\email{hyh422904725@gs.zzu.edu.cn}

\author{Zhenyu Yang}
\orcid{0009-0005-5298-0543}
\affiliation{
  \institution{Institute of Automation, CAS}
  \institution{UCAS}
  \city{Beijing}
  \country{China}
}
\email{yangzhenyu2022@ia.ac.cn}

\author{Shihan Wang}
\orcid{0009-0005-6218-7290}
\affiliation{
  \institution{Henan Institute of Advanced Technology, Zhengzhou University}
  \institution{Institute of Automation, CAS}
  \city{Beijing}
  \country{China}
}
\email{shihanwang@gs.zzu.edu.cn}

\author{Shengsheng Qian}
\orcid{0000-0001-9488-2208}
\authornote{Corresponding Author.}
\affiliation{%
  \institution{Institute of Automation, CAS}
  \institution{UCAS}
  \city{Beijing}
  \country{China}
  }
\email{shengsheng.qian@nlpr.ia.ac.cn}

\author{Bin Wen}
\orcid{0009-0006-2058-1609}
\affiliation{%
  \institution{Kuaishou Technology}
  \city{Beijing}
  \country{China}
}
\email{wenbin@kuaishou.com}

\author{Fan Yang}
\orcid{0009-0005-4570-5885}
\affiliation{
  \institution{Kuaishou Technology}
  \city{Beijing}
  \country{China}
}
\email{yangfan@kuaishou.com}

\author{Tingting Gao}
\orcid{}
\affiliation{
  \institution{Kuaishou Technology}
  \city{Beijing}
  \country{China}
}
\email{lisize@kuaishou.com}

\author{Changsheng Xu}
\orcid{0000-0001-8343-9665}
\affiliation{
  \institution{Institute of Automation, CAS}
  \institution{UCAS}
  \institution{Peng Cheng Laboratory}
    \city{Beijing}
    \country{China}
    }
\email{csxu@nlpr.ia.ac.cn}

\renewcommand{\shortauthors}{Yuhang Hu et al.}

\begin{abstract}
The rapid growth of streaming video applications demands multimodal models with enhanced capabilities for temporal dynamics understanding and complex reasoning.  However, current Video Question Answering (VideoQA) datasets suffer from two critical limitations: 1) Static annotation mechanisms fail to capture the evolving nature of answers in temporal video streams, and 2) The absence of explicit reasoning process annotations restricts model interpretability and logical deduction capabilities.  To address these challenges, We introduce \textbf{StreamingCoT}, the first dataset explicitly designed for temporally evolving reasoning in streaming VideoQA and multimodal Chain-of-Thought (CoT) tasks.  Our framework first establishes a dynamic hierarchical annotation architecture that generates per-second dense descriptions and constructs temporally-dependent semantic segments through similarity fusion, paired with question-answer sets constrained by temporal evolution patterns.  We further propose an explicit reasoning chain generation paradigm that extracts spatiotemporal objects via keyframe semantic alignment, derives object state transition-based reasoning paths using large language models, and ensures logical coherence through human-verified validation. This dataset establishes a foundation for advancing research in streaming video understanding, complex temporal reasoning, and multimodal inference. Our StreamingCoT and its construction toolkit can be accessed at \url{https://github.com/Fleeting-hyh/StreamingCoT}.
\end{abstract}



\maketitle

\section{Introduction}
The rapid proliferation of 5G networks and edge computing has propelled streaming video into a dominant medium for information dissemination, driving unprecedented demand for temporal-spatial reasoning in applications such as autonomous driving and intelligent surveillance. Multimodal models are now tasked with fine-grained parsing of dynamic spatiotemporal cues in video streams. Video Question Answering (VideoQA), a critical dataset for advancing multimodal cognitive capabilities, faces inherent limitations due to conventional datasets anchored in static timestamp-based annotations. While these datasets support rudimentary spatiotemporal reasoning, they inadequately address the dynamic evolution of answers in streaming scenarios, where answers may shift as events unfold continuously. This gap underscores a pivotal challenge: \textbf{how to design annotation frameworks that align with the fluid nature of streaming video to unlock models’ potential for real-world temporal reasoning?}

Existing VideoQA datasets predominantly employ static annotation mechanisms, where a single global answer is assigned to an entire video clip. Although efficient for annotation, this approach overlooks the continuity of event progression and the temporal dependencies inherent in answers. For instance, when objects undergo gradual state changes, static annotations fail to capture transitional cues, hindering models from discerning causal relationships or stage-specific attributes. This limitation points to \textbf{Challenge 1}: \textit{How to construct temporally sensitive question-answer pairs that reflect the dynamic, evolving nature of answers in streaming video contexts?}

Moreover, current datasets prioritize end-to-end answer prediction, providing only final labels without explicit annotations of intermediate reasoning steps. Consequently, models often exploit superficial statistical correlations between inputs and answers rather than deriving conclusions through multimodal reasoning. While chain-of-thought (CoT) methods have advanced textual reasoning tasks by modeling step-by-step logic, their adaptation to video modalities remains hindered by unaddressed complexities: (1) the continuous evolution of visual objects, (2) the hierarchical nature of temporal logic, and (3) the misalignment between multimodal semantics. These issues result in opaque decision-making processes, limiting model interpretability and multi-step reasoning fidelity. This leads to \textbf{Challenge 2}: \textit{How to establish an explicit reasoning chain annotation framework to enhance model transparency and logical coherence in streaming VideoQA?}

To address these challenges, we introduce \textbf{StreamingCoT}, the first streaming videoQA dataset featuring dynamic annotations and explicit reasoning chains. Our framework employs a \textit{hierarchical temporal annotation pipeline}: (1) \textbf{Dense per-second descriptions} are generated and fused into semantic units via similarity-based clustering, forming a temporally structured video narrative; (2) \textbf{Dynamically evolving question-answer pairs} are constructed under constrained templates, ensuring tight alignment between answer options and video content evolution. Additionally, we propose a \textbf{visualizable reasoning chain generation paradigm}: (1) Spatiotemporal objects are extracted via keyframe semantic alignment, (2) Large language models (LLMs) synthesize object state transitions into logical chains, and (3) Human validators refine these chains to ensure temporal consistency and coherence. This approach not only provides end-to-end learning signals but also enables traceable verification of reasoning processes.

Our contributions are threefold:
\vspace{-2mm}
\begin{itemize}
    \item \textbf{StreamingCoT Dataset}: A novel dataset for streaming VideoQA that replaces static annotations with a dynamic, temporally hierarchical labeling mechanism, addressing the rigidity of conventional datasets.
    \item \textbf{Explicit Reasoning Chain Framework}: A pioneering annotation methodology that deconstructs answers into interpretable spatiotemporal object state transitions, setting a new standard for evaluating model reasoning capabilities.
    \item \textbf{Standardized Pipeline}: A replicable workflow encompassing data collection, dynamic annotation, and multi-stage validation, offering methodological insights for building temporal-aware video understanding datasets.
\end{itemize}

\section{Related Work}

\subsection{VideoQA Dataset}
\label{sec:videoqa_dataset}

In the field of artificial intelligence, visual question answering (VQA) has been a research hotspot, while video question answering (VideoQA) is relatively under-researched compared to image question answering and has not been extensively explored ~\cite{khurana2021video}~\cite{tang2025video}. Traditional video question answering datasets, such as ~\cite{lei2018tvqa}~\cite{xiao2024can}~\cite{yu2019activitynet}, typically require annotators to watch videos, extract the core content, and generate corresponding question-answer pairs based on predefined standards. In recent years, new datasets and methods have emerged. For example, EgoSchema~\cite{mangalam2023egoschema} focuses on video understanding, particularly requiring the handling of longer video segments to answer complex questions. VideoCoT~\cite{wang2024videocot} and VCR-Bench~\cite{qi2025vcr} emphasize comprehensive reasoning over the entire video to generate more accurate question-answer results. SVBench~\cite{yang2025svbench}, as a contextual streaming video understanding dataset, enhances the accuracy of question answering by incorporating contextual information, but ignores explicit reasoning process annotations, which cannot model how conclusions are drawn from spatiotemporal dynamics. This paper proposes a dynamic hierarchical annotation framework that constructs temporally evolving QA pairs through similarity-guided semantic fusion and question type constraints, ensuring alignment with the dynamics of streaming videos.

\subsection{Visual CoT}
\label{sec:visual_cot}

Chain of Thought (CoT) reasoning can break down problem-solving into a series of continuous and interpretable steps~\cite{wang2025multimodal}~\cite{li2025perception}~\cite{xiao2024towards}, thereby assisting models in completing complex multi-step reasoning tasks. Early research~\cite{wei2022chain} indicated that CoT is an effective strategy for enhancing reasoning capabilities, and its effectiveness in the field of large language models (LLM) has also been widely validated~\cite{ma2023sci}. In the multimedia domain, studies such as ScienceQA~\cite{lu2022learn} and VisualCoT~\cite{rose2023visual} have emerged,while recent work in retrieval tasks, like SEIZE~\cite{yang2024semantic}and LDRE~\cite{yang2024ldre}, further demonstrates the versatility of structured reasoning in zero-shot composed image retrieval. Recently, VideoCoT~\cite{wang2024videocot} proposed an automatic annotation tool based on an active learning paradigm to generate video chain-of-thought datasets,but lacks a mechanism to align the reasoning path with the object state that evolves over time. The VoT~\cite{fei2024video} framework inherits the core idea of CoT, breaking down complex tasks into a series of simpler and more manageable sub-problems. CoS~\cite{hu2025cos}, on the other hand, focuses more on optimizing shot selection to enhance the model's understanding of long videos. This paper proposes a spatiotemporal-based CoT generation paradigm, aligning reasoning steps with key object trajectories and inter-frame state transitions. By utilizing vision large language model (VLLM) to synthesize human-in-the-loop verified object-centric chains of thought, this method ensures logical consistency and a visual temporal foundation, promoting interpretable reasoning in streaming video QA.

\begin{figure*}[t]
    \centering
    \includegraphics[width=\linewidth]{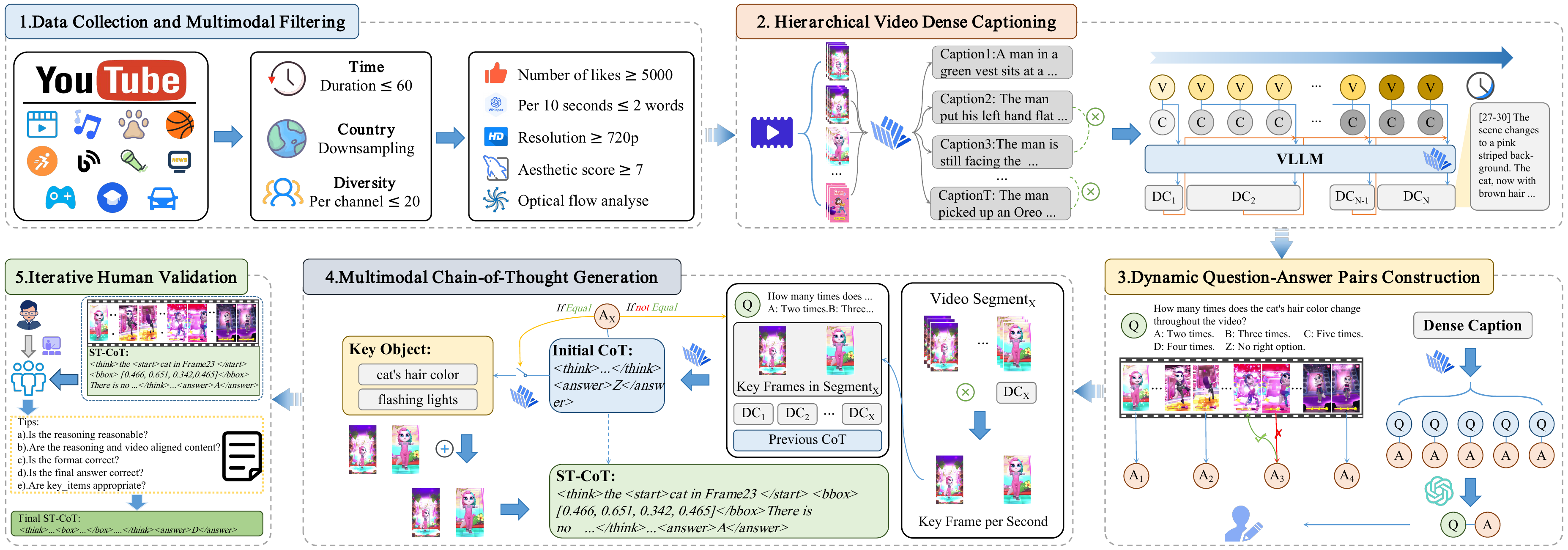}
    \caption{StreamingCoT Pipeline: Illustrates the hierarchical framework for dataset construction, comprising: (1) Geographically balanced video collection with multimodal filtering; (2) Adaptive temporal segmentation via Dynamic Semantic Fusion (DSF) and context-aware dense captioning; (3) Dynamic QA pair generation constrained by temporal evolution patterns, featuring distractor-aware option design; (4) Multimodal Chain-of-Thought synthesis integrating temporally verified reasoning, key object grounding, and spatiotemporal evidence fusion; (5) Iterative human validation ensuring spatiotemporal consistency and reasoning integrity throughout the workflow.}
    \label{fig:pipeline}
\end{figure*}

\section{Dataset: StreamingCoT}
\label{sec:dataset}

To advance research in streaming VideoQA and multimodal CoT reasoning, we introduce StreamingCoT, a large-scale dataset curated through a rigorous hierarchical pipeline that integrates temporal segmentation, dynamic question-answer generation, and multimodal evidence grounding. Below, we detail its construction, illustrated in Figure~\ref{fig:pipeline}, and key characteristics.

\vspace{-2mm}
\subsection{Data Collection and Multimodal Filtering}
\label{sec:data_filter}

\subsubsection{Data Collection}
\label{sec:data_collection}
We construct our video corpus through YouTube's official API, initially collecting 10,288 short-form videos with associated metadata. To mitigate geographic bias and enhance cross-cultural representation, we adopted a stratified sampling strategy based on video source countries. Two balancing mechanisms are implemented: (1) Region-based downsampling reduces overrepresentation from dominant geographic areas, and (2) Channel-level diversification limits video selection to 20 entries per channel to prevent content homogeneity. This dual approach ensures a global distribution and diversity in the dataset.

\subsubsection{Multimodal Filtering}
\label{sec:mul_filter}
To ensure high-quality cross-modal content, we design a three-stage hierarchical filtering architecture, progressively refining the dataset through social validation, audio content processing, and visual quality assessments.

\textbf{Social Validation.} 
We retain only videos demonstrating substantial audience engagement, filtering candidates using a threshold of $>$5,000 aggregated social interactions (likes, comments, shares). This crowd-sourced validation prioritizes content with verified community relevance.

\textbf{Audio Content Processing.}
Transcripts are generated using OpenAI's Whisper speech recognition system~\cite{Radford2022RobustSR}, with lexical density constraints applied to ensure content conciseness: transcripts exceeding two words per 10-second interval were discarded.

\textbf{Visual Quality Assessment.}
A multi-tiered screening pipeline is implemented:
(1) Resolution Screening: FFmpeg~\cite{tomar2006converting} preserves only HD content ($\geq$720p), removing low-resolution or heavily compressed videos.
(2) Motion Dynamics Analysis: Optical flow variance~\cite{2015opencv} identifies excessive motion blur ($\geq$15\% high-variance frames) or static segments ($\geq$85\% temporal occupancy in 30s windows)~\cite{yang2025svbench}, excluding videos with suboptimal temporal characteristics
(3) Aesthetic Evaluation: Like Open-Sora~\cite{Zheng2024OpenSoraDE}, we use the scoring model from CLIP+MLP Aesthetic Score Predictor to evaluate the aesthetic quality of videos. Videos scoring below 7/10 are filtered to maintain aesthetic integrity.

This iterative filtering reduced the initial 10,288 videos to 5,745 high-quality entries, balancing multimodal fidelity with content diversity.

\subsection{Hierarchical Video Dense Captioning}
\label{sec:dense_caption}

Our hierarchical video dense captioning framework addresses temporal redundancy and narrative coherence through a dual-stage architecture that integrates frame-level semantic analysis with dynamic temporal segmentation. For each video $V$, we first generate fine-grained text captions \{$C_{1},...,C_{n}$\} at 1-second intervals based on InternVL3~\cite{chen2024expanding}, thereby establishing a frame-level semantic baseline with precise temporal alignment.

\subsubsection{Adaptive Temporal Segmentation}
\label{sec:temporal_seg}

The Dynamic Semantic Fusion (DSF) algorithm employs a Last-In-First-Out(LIFO) stack architecture with thresholding ($\theta = 0.9$)~\cite{Han2024VideoEspressoAL}. For timestamp $t$, we compute cosine similarity $S_{t-1,t}$ between consecutive captions $C_{t-1}$ and $C_t$:

\begin{equation}
    S_{t-1,t} = \frac{\mathbf{E}(C_{t-1}) \cdot \mathbf{E}(C_t)}{\|\mathbf{E}(C_{t-1})\| \|\mathbf{E}(C_t)\|}
\end{equation}

where $\mathbf{E}(\cdot)$ denotes the sentence embedding function. For each segment ${Seg}_i$, the segmentation rule is formalized as:

\begin{equation}
    {Seg}_i = 
    \begin{cases} 
        \text{Merge}(t_{k-n},...,t_k) & \text{if } \prod_{m=k-n}^{k-1} S_{m,m+1} \geq \theta \\
        \text{Push}(t_k) & \text{otherwise}
    \end{cases}
\end{equation}

\subsubsection{Context-Aware Narration Generation}
\label{sec:context_generation}

The narration synthesis combines intra-segment compression and inter-segment contextualization. To prevent cumulative delays in subsequent subtitles, the 0-1 second caption segment is anchored and remains distinct from following captions without integration. For segment ${Seg}_i$:

\begin{equation}
    {DC}_i = \text{VLLM}_{\text{merge}}(C_{t_s},...,C_{t_e}; \mathcal{W}_{\text{visual}}; {DC}_{1},..., {DC}_{i-1})
\end{equation}

where $DC$ denotes dense caption, and $\mathcal{W}_{\text{visual}}$ denotes the visual features of the video segments. Inter-segment coherence is maintained through the historical $DC$.

\subsubsection{Quality Validation}
\label{sec:quality_validation}

To rigorously evaluate the quality and coherence of the generated Dense Captions (DCs), we conducted a comprehensive human annotation study following established best practices for video description assessment. A panel of twenty domain experts, each possessing advanced qualifications in linguistics or computer vision and undergoing standardized task-specific training, independently annotated these videos. 
Annotation guidelines emphasized evaluating three critical dimensions:
(1)Semantic Completeness \& Accuracy: Verifying that all salient visual objects, actions, relationships, and scene context described in the DC were factually present and correctly interpreted within the corresponding $Seg_i$.
(2)Narrative Coherence: Assessing the logical flow and temporal consistency of the DC within its segment  and its contextual linkage to preceding segments via the historical DC context, ensuring smooth transitions and avoidance of contradictory statements.
(3)Temporal Alignment: Confirming the precise synchronization between the generated DC and the visual events depicted in the video segment, particularly scrutinizing the handling of anchored segments and merged intervals to detect temporal misalignment or delay propagation.

\subsection{Dynamic Question-Answer Pairs Construction}
\label{sec:qa_construction}

To address the temporal evolution of answers in streaming video contexts, we propose a systematic framework for generating dynamic question-answer (QA) pairs that explicitly adapt to shifting semantic states across video timelines. Our methodology comprises four key stages: Dynamic QA Generation, Distractor-Aware Option Design, Two-Stage Quality Quality Filtering, and Temporal Segment Realignment, as illustrated in Figure~\ref{fig:pipeline}.

\subsubsection{Dynamic QA Generation} 
\label{sec:qa_generation}

Leveraging InternVL2.5's multimodal reasoning capabilities, we generate candidate QA pairs by integrating dense captions with raw video content. Each video yields five initial QA pairs, constrained to six question types:
(1) Cumulative counting: Counting the total number of times a certain type of activity, phenomenon or object appears in the entire video, which requires continuous tracking rather than single-frame judgment.
(2) Periodic pattern recognition: Inferring patterns through recurring patterns.
(3) Sequential step recognition: Identifying the different stages of a process or action in a video.
(4) State duration: Determining the duration of a state or action in a video.
(5) Object state recognition: Identifying and tracking state changes or attribute changes of an object based on the video screen.
(6) Clue-revealing response: In a video, responding is not made until sufficient clues or information appear.

To refine candidate quality, we implement a scoring mechanism using GPT-4o's multimodal evaluation capabilities. Each of the five generated QA pairs per video undergoes quantitative assessment based on three criteria: (1) Visual-grounding consistency between question semantics and video content, (2) Temporal validity of answers relative to event dynamics, and (3) Reasoning complexity alignment with human cognition. The model computes a normalized quality score through weighted aggregation of these metrics, with the top-ranked QA pair selected for downstream processing. 

\subsubsection{Distractor-Aware Option Design}
\label{sec:distractor_design}

We engineer semantically plausible distractors that exploit characteristic temporal misperceptions via InternVL3. For each generated question-answer (QA) pair, we synthesize three distractors through systematic perturbation of temporal dependencies, ensuring they reflect common reasoning failures in video understanding. Distractor construction adheres to the following principles:
1)Temporal Misalignment: Distractors incorporate elements from adjacent but semantically distinct video segments.
2)Partial Pattern Compliance: For periodic or cumulative queries, distractors satisfy subsets of temporal conditions while violating critical global constraints.
3)State Transition Fallacies: In object-state recognition tasks, distractors represent physically plausible yet temporally inconsistent state transitions derived from invalid spatiotemporal pathways.
4)Premature Inference: For clue-revealing responses, distractors reflect conclusions drawn before sufficient evidence accumulation.

\subsubsection{Human Verification} 
\label{sec:human_verify}

To ensure temporal grounding fidelity and semantic accuracy in the constructed QA pairs, we implement a rigorous human verification protocol. Expert annotators meticulously evaluate each selected QA pair against the corresponding video timeline, focusing on two critical dimensions: temporal consistency and answer validity.

Annotations are performed at the segment-granular level, where videos are partitioned into semantically coherent intervals derived from the original dense captions. For each segment, annotators assess:
1) Answer Temporal Alignment: Verification that ground-truth answers correspond precisely to relevant video segments, including confirmation that:
(a) Cumulative counts reflect events occurring only within the relevant temporal scope
(b) State durations are measured across contiguous segments satisfying described conditions
(c) Sequential steps follow actual chronological ordering
(d) Clue-revealing answers emerge after sufficient evidence manifests
2) Distractor Temporal Plausibility: Validation that synthetically generated distractors exhibit legitimate temporal misalignments while remaining factually incorrect
3) Segment-Answer Coherence: Certification that answers remain invariant when considering only essential temporal segments, excluding irrelevant periods.

\subsection{Multimodal Chain-of-Thought Generation}
\label{sec:cot_generation}

Our framework generates spatiotemporally grounded reasoning chains through a multi-stage process that integrates temporal dynamics and spatial context. The methodology formalizes three core computational stages:

\subsubsection{Temporally-Aware CoT Initialization}
\label{sec:initial_cot}

For video segment $Seg_i$, we identify the keyframe \(Keyframe_i\) through frame-dense caption similarity maximization:
\begin{equation}
    Keyframe_i = \{\underset{f_i \in \mathcal{F}_t}{\arg\max}  \ \text{Sim}\big(\mathbf{E}_{\text{vis}}(f_i),  \mathbf{E}_{\text{text}}(DC_i)\big)\mid t \in Seg_i\}
\end{equation}
where \(\mathcal{F}_t\) denotes $t$ second in \(Seg_i\), \(DC_t\) the dense caption, and \(\mathbf{E}\) visual/text encoders. 

The initial CoT \(CoT_i^{\text{init}}\) is then generated via InternVL-2.5 by conditioning on: (1) the current segment's dense caption, (2) keyframe-selected historical dense captions, and (3) preceding CoT sequences. This ensures temporal continuity across segments.
\begin{equation}
    CoT_i^{\text{init}} = {\text{VLLM}}\big(DC_i,  \{ DC_{i-\tau}, CoT_{i-\tau} \}_{\tau=1}^{H},  Keyframe_i \big)
\end{equation}
with \(\text{VLLM}\) as vision large language model and \(H\) the historical context window. Crucially, we perform temporally constrained verification: the answer derived from the initial CoT is rigorously compared against the segment's ground truth. In cases of inconsistency, the CoT undergoes regeneration to enforce factual alignment with evolving video content.

\subsubsection{Spatiotemporal key Object Grounding}
\label{sec:key_object}

We extract critical key objects $Obj_i = \{o_x \mid x \leq 3\}$ referenced in \(CoT_t^{\text{init}}\) using InternVL2.5's semantic parsing capabilities.
\begin{equation}
    Obj_i = {\text{VLLM}}(CoT_t^{\text{init}})
\end{equation}]
These key objects $Obj_i$ are then spatially grounded within their $Keyframe_i$ using GroundingDINO, which generates precise Bounding Boxes (BBoxes).
\begin{equation}
    BBoxs_i = \left\{ (o_i, \text{GroundingDINO}(o_i, Keyframe_i))  \mid  \forall o_i \in Obj_i \right\}
\end{equation}
where \(BBoxs_i\) represents bounding boxes capturing spatial coordinates.

\subsubsection{Multimodal Reasoning Fusion}
\label{sec:reason_fusion}

The final Spatiotemporal-CoT \(CoT_t^{\text{ST}}\) is synthesized seamlessly by integrating three complementary elements: (1) the temporally verified reasoning steps of the initial CoT, (2) the temporal anchors provided by keyframe timestamps, and (3) the spatial evidence encoded by the bounding boxes of key objects.
\begin{equation}
    CoT_t^{\text{ST}} = \Phi_{\text{fuse}}\big( CoT_t^{\text{init}},  Keyframe_i,  BBoxs_i \big)
\end{equation}
This fusion creates an auditable reasoning trace where each deduction step satisfies the spatiotemporal grounding condition:
\begin{equation}
    \forall r_j \in CoT_t^{\text{ST}} \ \exists \ (o_j, t_j, bbox_j) : r_j \propto \mathcal{V}(o_j, t_j) \otimes \mathcal{S}(bbox_j)
\end{equation}
with \(r_j\) denoting reasoning steps, \(\mathcal{V}\) visual features, \(\mathcal{S}\) spatial features, and \(\otimes\) the multimodal fusion operator. The resultant \(\mathbf{C}_t^{\text{ST}}\) provides explicit alignment between logical inferences and their multimodal evidence sources.

\subsection{Iterative Human Validation}
\label{sec:iterative_human}

To ensure the spatiotemporal fidelity and reasoning integrity of the generated Chains-of-Thought (CoT\(_t^{\text{ST}}\)), we implement a rigorous three-phase iterative validation protocol conducted by domain specialists. This process guarantees that every reasoning chain satisfies strict multimodal grounding criteria before final inclusion in the dataset.

\subsubsection{Validation Protocol}
\label{sec:validation_protocol}

A panel of 15 certified annotators (with advanced degrees in computer vision or computational linguistics) undergoes task-specific training using standardized guidelines. Each \(CoT_t^{\text{ST}}\) undergoes independent evaluation by one annotator focusing on four critical dimensions:
1) Spatiotemporal Consistency: Verification that all referenced objects ($Obj_i$) maintain correct spatial coordinates ($BBoxs_i$) throughout their temporal appearance windows, with strict enforcement of the grounding condition: 
\[
\forall (o_j, t_j, bbox_j) \in CoT_t^{\text{ST}}, \quad \mathcal{V}(o_j, t_j) \otimes \mathcal{S}(bbox_j) \neq \emptyset
\]  
2) Temporal Causality: Audit of logical transitions between reasoning steps $\{r_j\}$ to ensure strict adherence to chronological event ordering and elimination of anachronistic inferences. 
3) Evidence Completeness: Certification that each deductive step $r_j$ explicitly references sufficient visual evidence from $Keyframe_i$ or historically linked segments $\{DC_{i-\tau}\}_{\tau=1}^H$.   
4) Answer Derivation Soundness: Validation that the final answer emerges conclusively from the reasoning trace without logical gaps or external knowledge contamination.

\subsubsection{Iterative Refinement Mechanism}
\label{sec:iterative}

Annotators flag deficient CoTs using a structured taxonomy of error types. Rejected cases trigger a \textit{corrective regeneration pipeline}:

\begin{equation}
\footnotesize
CoT_t^{\text{ST*}} = \Phi_{\text{fuse}}\Big( \underbrace{\text{VLLM}_{\text{revise}}(CoT_t^{\text{init}}, \mathcal{E}_{\text{annot}})}_{\text{Revised Reasoning}}, Keyframe_i, \underbrace{\text{ReGround}(Obj_i)}_{\text{Updated BBoxs}} \Big)
\end{equation}

where $\mathcal{E}_{\text{annot}}$ denotes annotator feedback. Each regenerated CoT undergoes re-validation, with maximum three iterations per sample. Samples that fail validation after the last iteration are manually annotated.

\begin{figure}[t]
    \centering
    \includegraphics[width=\linewidth]{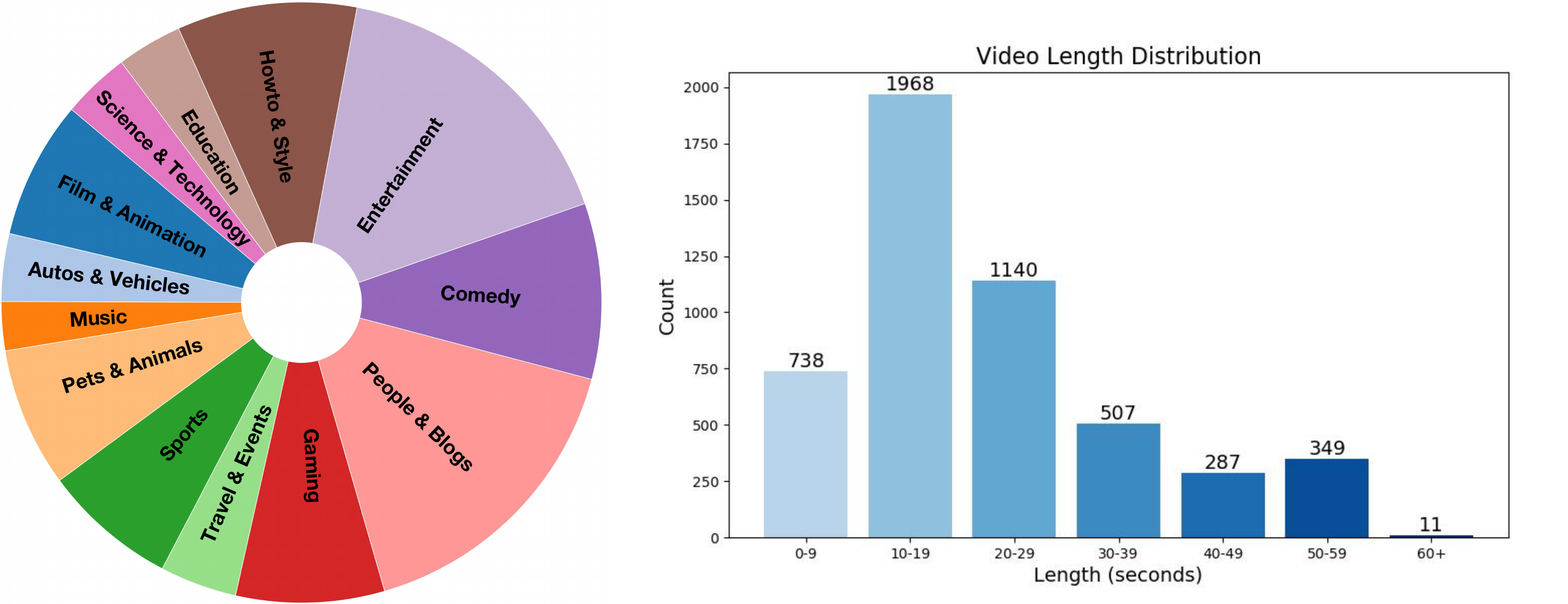}
    \caption{Dataset Statistics}
    \label{fig:dataset_statistics}
\end{figure}

\begin{figure*}[ht]
    \centering
    \includegraphics[width=0.95\linewidth]{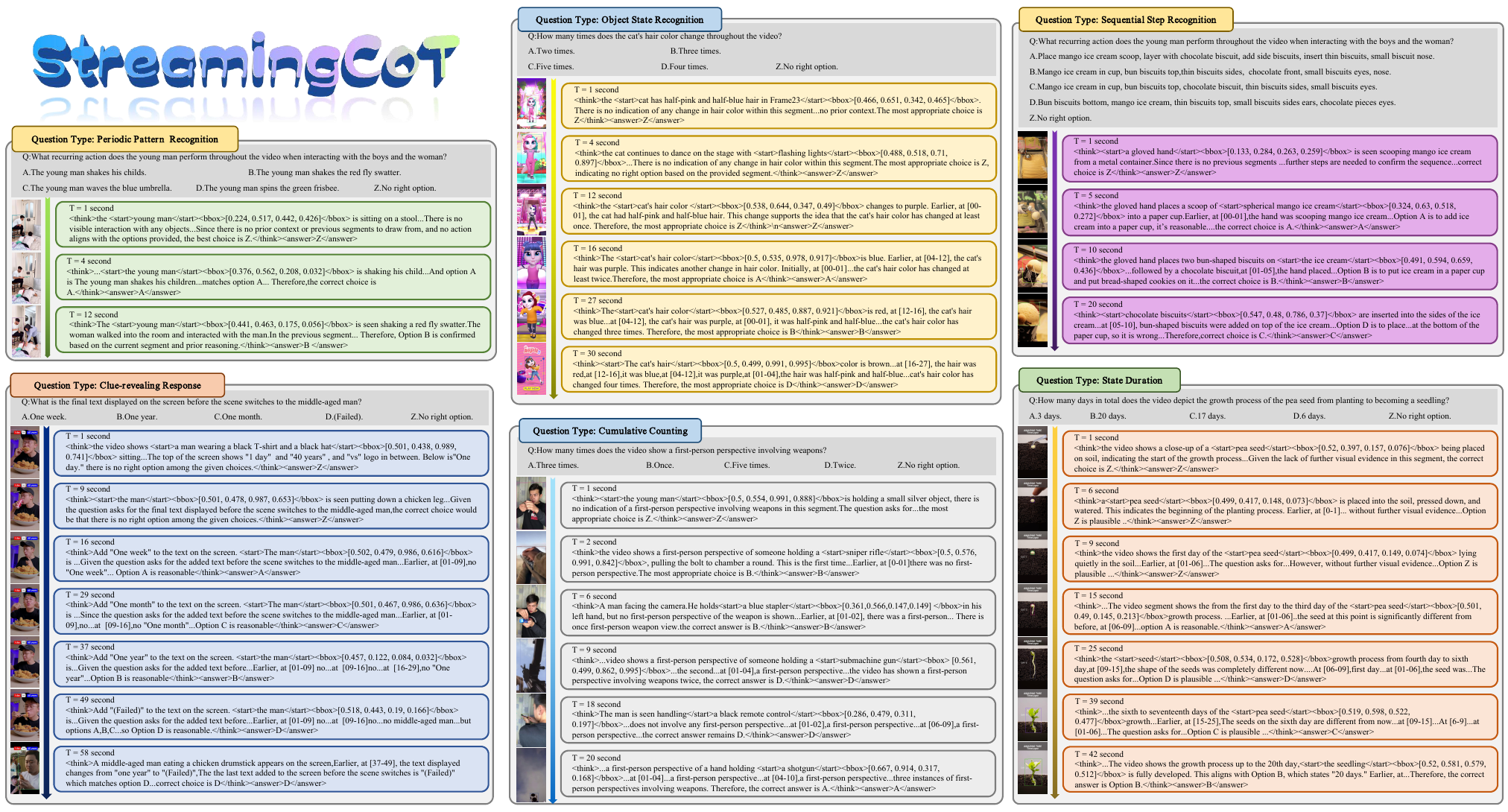}
    \caption{Examples of question types and temporal evidence accumulation in the StreamingCoT dataset}
    \label{fig:Example}
\end{figure*}

\subsection{Dataset Statistics}
\label{sec:dataset_statistics}

As shown in Figure~\ref{fig:dataset_statistics}, StreamingCoT comprises 5,000 high-quality short-form videos meticulously filtered through multimodal quality assessments, with an average duration of 25.6 seconds. The dataset combines 243,185 time-anchored dense subtitles per second into 68,940 semantic segments, with an average of 12 segments per video, using our dynamic semantic fusion algorithm. These segments form the foundation for 34,470 dynamic question-answer pairs (5 QA pairs per video) spanning six temporally evolving question types: cumulative counting (18.2\%), periodic pattern recognition (15.7\%), sequential step recognition (22.4\%), state duration (16.1\%), object state recognition (19.3\%), and clue-revealing responses (8.3\%). Each QA pair includes three human-verified distractors engineered to reflect characteristic temporal reasoning failures. Crucially, the dataset features 68,940 multimodal Chain-of-Thought (CoT) annotations with spatiotemporal grounding, incorporating 206,820 key object bounding boxes localized through our keyframe alignment protocol. StreamingCoT covers 32 thematic categories including instructional activities (24.1\%), natural phenomena (18.7\%), social interactions (22.3\%), mechanical processes (17.9\%), and artistic performances (16.9\%). 

\section{Case Study}
\label{sec:case_study}

This case study demonstrates the dynamic nature of streaming visual question answering, where answers evolve over time and require contextual reasoning to answer, requiring spatiotemporal reasoning to capture complex patterns.

\textbf{Temporal context integration.}
\label{sec:case_study_context}
Requires integrating temporal clues to resolve ambiguity, As shown in Figure~\ref{fig:Example},in the young man example. Early segments lack sufficient context, producing indeterminate answers until later timestamps reveal key interactions. Similarly, the cat’s hair color question initially lacks decisive evidence, but cumulative observations across segments expose the correct answer through progressive state changes. This demonstrates how streaming QA relies on temporal evidence accumulation to overcome early uncertainty.

\textbf{Dynamic Answer Adaptation.}
\label{sec:answer_adaptation}
Answers evolve dynamically as new information arrives. The text-addition example shows responses shifting from provisional to final as overlays appear, mirroring real-world scenarios where answers depend on temporal closure. The pea seed’s growth duration also refines over time, starting vague before converging to the correct value. These cases highlight the need for models to update hypotheses incrementally rather than relying on static snapshots.

\textbf{Spatiotemporal Reasoning.}
\label{spatiotemporal_reason}
Sequential tasks like ice cream assembly demand both spatial and temporal reasoning. Correct answers emerge only after tracking object placements in order, such as layering biscuits after scooping ice cream. The weapon-counting task similarly requires identifying spatial details while accumulating instances over time. These examples illustrate how streaming QA combines spatial grounding with temporal sequencing for accurate reasoning.

\section{Conclusions}
\label{sec:conclusions}

This paper introduces \textbf{StreamingCoT}, a foundational dataset designed to address critical limitations in streaming VideoQA and multimodal reasoning research. 
StreamingCoT's distinctive value lies in its integration of temporally evolving answer spaces with explicit multimodal reasoning traces—features absent in existing benchmarks. The dataset establishes new standards for evaluating temporal understanding capabilities in multimodal systems while providing auditable reasoning pathways that enhance model interpretability. 

\newpage
\begin{acks}
This work was supported by the National Key Research and Development Program of China (No.2023YFC3310700), the Beijing Natural Science Foundation (JQ23018) and the National Natural Science Foundation of China (No. 62036012, 62276257).
\end{acks}

\bibliographystyle{ACM-Reference-Format}
\balance
\bibliography{sample-base}
\end{document}